# Identifying Independencies in Causal Graphs with Feedback

**Judea Pearl**
Cognitive Systems Laboratory
Computer Science Department
University of California, Los Angeles, CA 90024
*judea@cs.ucla.edu*

**Rina Dechter**
Information & Computer Science
University of California, Irvine
Irvine, CA 92717
*rdechter@ics.uci.edu*

## Abstract

We show that the $d$-separation criterion constitutes a valid test for conditional independence relationships that are induced by feedback systems involving discrete variables.

## 1   INTRODUCTION

It is well known that the $d$-separation test is sound and complete relative to the independencies assumed in the construction of Bayesian networks [Verma and Pearl, 1988, Geiger et al., 1990]. In other words, any $d$-separation condition in the network corresponds to a genuine independence condition in the underlying probability distribution and, conversely, every $d$-connection corresponds to a dependency in at least one distribution compatible with the network.

The situation with feedback systems is more complicated, primarily because the probability distributions associated with such systems do not lend themselves to a simple product decomposition. The joint distribution of feedback systems cannot be written as a product of the conditional distributions of each child variable, given its parents. Rather, the joint distribution is governed by the functional relationships that tie the variables together.

Spirtes (1994) and Koster (1995) have nevertheless shown that the $d$-separation test is valid for cyclic graphs, provided that the equations are linear and all distributions are Gaussian. In this paper we extend the results of Spirtes and Koster and show that the $d$-separation test is valid for nonlinear feedback systems and non-Gaussian distributions, provided the variables are discrete.

## 2   BAYESIAN NETWORKS VS. CAUSAL NETWORKS: A REVIEW

In this section we first review the basic notions and nomenclature associated with Bayesian networks, and then we contrast these notions with those associated with recursive and nonrecursive causal models. The reader is encouraged to consult the example in section 3.1, where these notions are given graphical representations.

### 2.1   BAYESIAN NETWORKS

**Definition 1** *Let $V = \{X_1, \ldots, X_n\}$ be an ordered set of variables, and let $P(v)$ be the joint probability distribution on these variables. A set of variables $PA_j$ is said to be Markovian parents of $X_j$ if $PA_j$ is a minimal set of predecessors of $X_i$ that renders $X_j$ independent of all its other predecessors. In other words, $PA_j$ is any subset of $\{X_1, \ldots, X_{j-1}\}$ satisfying*

$$P(x_j|pa_j) = P(x_j|x_1, \ldots, x_{j-1}) \qquad (1)$$

*such that no proper subset of $PA_j$ satisfies Eq. (1).*

Definition 1 assigns to each variable $X_j$ a select set of variables $PA_j$ that are sufficient for determining the probability of $X_j$; knowing the values of other preceding variables is redundant once we know the values $pa_j$ of the parent set $PA_j$. This assignment can be represented in the form of a directed acyclic graph (DAG) in which variables are represented by nodes and arrows are drawn from each node of the parent set $PA_j$ toward the child node $X_j$. Definition 1 also suggests a simple recursive method for constructing such a DAG: At the $i$th stage of the construction, select any minimal set of $X_i$'s predecessors that satisfies Eq. (1), call this set $PA_i$ (connoting "parents"), and draw an arrow from each member in $PA_i$ to $X_i$. The result is a DAG, called a *Bayesian network* [Pearl, 1988], in which an arrow from $X_i$ to $X_j$ assigns $X_i$ as a Markovian parent of $X_j$, consistent with Definition 1.

The construction implied by Definition 1 defines a Bayesian network as a carrier of conditional independence information that is obtained along a specific order $O$. Clearly, every distribution satisfying Eq. (1) must decompose (using the chain rule of probability calculus) into the product

$$P(x_1, \ldots, x_n) = \prod_i P(x_i \mid pa_i) \qquad (2)$$

which is no longer order-specific. Conversely, for every distribution decomposed as Eq. (2) one can find an ordering $O$ that would produce $G$ as a Bayesian network. If a probability distribution $P$ admits the product decomposition dictated by $G$, as given in Eq.(2), we say that $G$ and $P$ are *compatible*.

A convenient way of characterizing the set of distributions compatible with a DAG $G$ is to list the set of (conditional) independencies that each such distribution must satisfy. These independencies can be read off the DAG by using a graphical criterion called *d-separation* [Pearl, 1988]. To test whether $X$ is independent of $Y$ given $Z$ in the distributions represented by $G$, we need to examine $G$ and test whether the nodes corresponding to variables $Z$ d-separate all paths from nodes in $X$ to nodes in $Y$. By *path* we mean a sequence of consecutive edges (of any directionality) in the DAG.

**Definition 2** (*d*-separation) *A path p is said to be d-separated (or blocked) by a set of nodes Z iff:*

(i) *p contains a chain $i \longrightarrow j \longrightarrow k$ or a fork $i \longleftarrow j \longrightarrow k$ such that the middle node $j$ is in $Z$, or*

(ii) *p contains an inverted fork $i \longrightarrow j \longleftarrow k$ such that neither the middle node $j$ nor any of its descendants (in $G$) are in $Z$.*

*If $X, Y$, and $Z$ are three disjoint subsets of nodes in a DAG $G$, then $Z$ is said to d-separate $X$ from $Y$, denoted $(X \underline{\parallel} Y|Z)_G$, iff $Z$ d-separates every path from a node in $X$ to a node in $Y$.*

To distinguish between the graphical notion of $d$-separation, $(X \underline{\parallel} Y|Z)_G$, and the probabilistic notion of conditional independence, we will use the notation $(X \perp Y|Z)_P$ for the latter. The connection between the two is given in Theorem 1.

**Theorem 1**
[Verma and Pearl, 1988, Geiger et al., 1990] *For any three disjoint subsets of nodes $(X, Y, Z)$ in a DAG $G$, and for all probability functions $P$, we have*

(i) $(X \underline{\parallel} Y|Z)_G \Longrightarrow (X \perp Y|Z)_P$ *whenever $G$ and $P$ are compatible, and*

(ii) *If $(X \perp Y|Z)_P$ holds in all distributions compatible with $G$, then $(X \underline{\parallel} Y|Z)_G$.*

An alternative test for $d$-separation has been devised by Lauritzen et al. (1990), based on the notion of moralized ancestral graphs. To test for $(X \underline{\parallel} Y|Z)_G$, delete from $G$ all nodes except those in $\{\overline{X}, Y, Z\}$ and their ancestors, connect by an edge every pair of nodes that share a common child, and remove all arrows from the arcs. $(X \underline{\parallel} Y|Z)_G$ holds iff $Z$ is a cutset of the resulting undirected graph, separating nodes of $X$ from those of $Y$.

This alternative test of $d$-separation will play a major role in our proofs, hence we cast it in some extra notation. Denote by $(X \underline{\parallel}^* Y|Z)_G$ the condition that $Z$ intercepts all paths between $X$ and $Y$ in some undirected graph $G$. Let $G^{XYZ}$ stand for the undirected ancestral graph obtained through Lauritzen's construction. Lauritzen's equivalence theorem amounts to asserting

$$(X \underline{\parallel}^* Y|Z)_{G^{XYZ}} \Longleftrightarrow (G \underline{\parallel} Y|Z)_G \qquad (3)$$

Note that the operator $\underline{\parallel}^*$ denotes ordinary separation in undirected graphs while $\underline{\parallel}$ stands for $d$-separation in DAGs.

## 2.2 CAUSAL THEORIES AND CAUSAL GRAPHS

A causal theory is a fully specified model of the causal relationships that govern a given domain, that is, a mathematical object that provides an interpretation (and computation) of every causal query about the domain. Following [Pearl, 1995b] we will adapt here a definition that generalizes most causal models used in engineering and economics.

**Definition 3** *A causal theory is a four-tuple*

$$T = < V, U, P(u), \{f_i\} >$$

*where*

(i) $V = \{X_1, \ldots, X_n\}$ *is a set of observed variables,*

(ii) $U = \{U_1, \ldots, U_n\}$ *is a set of exogenous (often unmeasured) variables that represent disturbances, abnormalities, or assumptions,*

(iii) $P(u)$ *is a distribution function over $U_1, \ldots, U_n$, and*

(iv) $\{f_i\}$ *is a set of $n$ deterministic functions, each of the form*

$$x_i = f_i(pa_i, u_i) \quad i = 1, \ldots, n \qquad (4)$$

*where $PA_i$ is a subset of variables in $V$ not containing $X_i$, and $pa_i$ is any instance of $PA_i$.*

*We will further assume that the set of equations in (iv) has a unique solution for $X_1, \ldots, X_n$, given any value of the disturbances $U_1, \ldots, U_n$.[1] Therefore, the distribution $P(u)$ induces a unique distribution on the observables, which we denote by $P_T(v)$.*

As in the case of Bayesian networks, drawing arrows between the variables $PA_i$ and $X_i$ defines a directed graph $G_T$, which we call the *causal graph* of $T$. However, the construction of causal graphs differs from that of Bayesian networks in two ways. First, $G_T$ may, in general, be cyclic. Second, unlike the Markovian parents in Definition (1), $PA_i$ are specified by the modeler based on assumed functional relationships among the variables, not by conditional independence considerations, as in Bayesian networks.[2]

---

[1]The uniqueness assumption is equivalent to the requirement that the set of equations $\{f_i\}$, viewed as a dynamic physical system, be stable. Indeed, if $V$ may attain two different states under the same $U$, it means that a small perturbation in $U$ would result in a drastic change in $V$, hence, instability.

There is a special class of causal theories, called *Markovian*, where the two specification schemes coincide.

**Definition 4** *A causal theory is said to be Markovian if two conditions are satisfied:*

*(i) the theory is recursive, that is, there exists an ordering of the variables $V = \{X_1, \ldots, X_n\}$ such that each $X_i$ is a function of a subset $PA_i$ of its predecessors*

$$x_i = f_i(pa_i, u_i) \qquad PA_i \subseteq \{X_1, \ldots, X_{i-1}\} \quad (5)$$

*(ii) the disturbances $U_1, \ldots, U_n$ are mutually independent, that is, for all $i$*

$$U_i \perp\!\!\!\perp \{U \setminus U_i\} \quad (6)$$

It is easy to see (e.g., [Pearl and Verma, 1991]) that the distribution induced by any Markovian theory $T$ is given by the product in Eq. (2),

$$P_T(x_1, \ldots, x_n) = \prod_i P_T(x_i | pa_i) \quad (7)$$

where $pa_i$ are the parents of $X_i$ in the causal graph of $T$. Hence, the causal graphs associated with Markovian theories coincide with the Bayesian networks induced by those theories. In general, however, the causal graphs associated with non-Markovian theories may be cyclic, and the set of independencies induced by such theories would not be represented by a DAG. The purpose of this paper is to show that those independencies nevertheless can be read off the causal graph using the $d$-separation test, provided two conditions are satisfied: (1) Eq. (6) holds, and (2) the variables in $V$ are discrete.

---

[2] Definition 3 is a generalization of "structural equations" in econometrics [Goldberger, 1972], where continuous variables are normally assumed, and where the analysis is generally confined to linear systems with Gaussian noise. It should be emphasized that a set of equations as the one described in Definition 3, although sufficient for the purposes of this paper, would not, in itself, warrant the title "causal theory". Notions of autonomy and interventions should be an integral part of any such definition because the primary function of causal theories, setting them apart from algebraic equations or regression models, is to predict the effects of unanticipated changes, e.g., external interventions not modeled in $U$. Each such intervention corresponds to modifying a select set of equations while keeping the others intact (see Appendix 1 in Pearl (1995a), or Pearl (1995b).

Causal theories that obey Eq. (6) but possibly not Eq. (5) will be called *semi-Markovian*. Such theories are "complete" in the sense that all probabilistic dependencies are explained in terms of causal dependencies. We will first state our results for semi-Markovian theories and then extend them to general, non-Markovian theories.

## 3 THE MAIN RESULTS

Our main result is stated in the following theorem.

**Theorem 2** *Given a semi-Markovian causal theory $T$, with an associated causal graph $G_T$, if each variable in $V$ has a discrete and finite domain, then*

$$(X \perp\!\!\!\perp Y | Z)_{G_T} \Longrightarrow (X \perp Y | Z)_{P_T} \quad (8)$$

The restriction that $T$ be semi-Markovian is not a severe one. Theorem 2 can be extended to general, non-Markovian theories through the notion of an *augmented* graph.

**Definition 5** (augmented graph) *Given a causal theory $T$ with an associated causal graph $G_T$, the augmented graph $G'_T$ of $T$ is a graph constructed by adding a set $D$ of dummy root nodes to $G_T$, where each dummy node points at two $U$ nodes, such that any two dependent subsets of $U$ variables (i.e., $U_1$ and $U_2$ such that $P(u_1, u_2) \neq P(u_1)P(u_2)$) will have a common ancestor in $D$.*

In the literature on path analysis [Wright, 1921] and structural equation models [Goldberger, 1972], it is common to designate these dummy nodes by curved, bidirected arcs connecting two disturbances. They represent unobserved common factors that the modeler has decided to keep outside the analysis.

**Corollary 1** *Given a general causal theory $T$ with an associated causal graph $G_T$, if each variable in $V$ has a discrete and finite domain, then*

$$(X \perp\!\!\!\perp Y | Z)_{G'_T} \Longrightarrow (X \perp Y | Z)_{P_T} \quad (9)$$

where $G'_T$ is an augmented graph of $T$.

### 3.1 EXAMPLE

Consider the cyclic graph $G_T$ shown in Figure 1, which represents a semi-Markovian causal theory given by the following four equations

$$\begin{aligned} x_1 &= f_1(u_1) \\ x_2 &= f_2(u_2) \\ x_3 &= f_3(x_1, x_4, u_3) \\ x_4 &= f_4(x_2, x_3, u_4) \end{aligned} \quad (10)$$

The disturbances $(U_1, U_2, U_3, U_4)$ are not part of $G_T$, and are shown here for clarity, using dashed arrows.

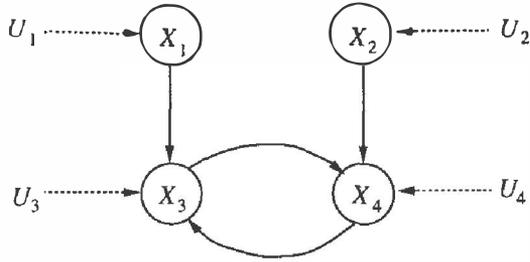

Figure 1: A cyclic graph associated with the causal theory of (10).

The augmented graph $G'_T$ is identical to $G_T$, because the $U$'s are assumed mutually independent. (Had any two of the $U$'s been dependent, say $U_1$ and $U_2$, the augmented graph $G'_T$ would contain a dummy node as a common parent of $X_1$ and $X_2$.)

$G_T$ advertises two $d$-separation conditions

$$(X_1 \;\|\; X_2|\; \emptyset)_{G_T} \qquad (11)$$
$$(X_1 \;\|\; X_2|\; \{X_3, X_4\})_{G_T} \qquad (12)$$

These can be verified either by direct application of Definition 2, or by constructing the corresponding moralized ancestral graphs (shown in Figure 2) and noting that the graph separation conditions: $(X_1 \;\|^*\; X_2)_{G^{X_1 X_2}}$ and $(X_1 \;\|^*\; X_2|X_3 X_4)_{G^{X_1 X_2 X_3 X_4}}$ are valid.

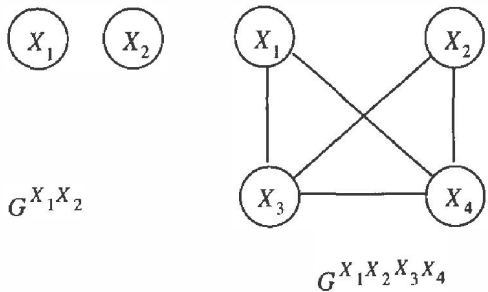

Figure 2: Moralized ancestral graphs corresponding to Eqs. (11) and (12).

The two separations advertised by $G_T$, Eqs. (11)–(12), represent two independence claims

$$(X_1 \;\perp\; X_2)_{P_T} \qquad (13)$$
$$(X_1 \;\perp\; X_2|X_3 X_3)_{P_T} \qquad (14)$$

about the distribution $P_T(x_1, x_2, x_3, x_4)$ induced by $T$. Theorem 2 ensures that these claims are valid regardless of the functions $(f_1, f_2, f_3, f_4)$ or the distributions $\{P(u_1), P(u_2), P(u_3), P(u_4)\}$, as long as $X_1, X_2, X_3, X_4$ are discrete and equations (10) have a unique solution for $(x_1, x_2, x_3, x_4)$.

We will exemplify Theorem 2 through a specific causal theory satisfying the discreteness and uniqueness conditions. Consider the theory:

$$\begin{aligned}
x_1 &= u_1 \\
x_2 &= u_2 \\
x_3 &= \begin{cases} g_0(x_4) & \text{if } x_1 \vee u_3 = 0 \\ g_1(x_4) & \text{if } x_1 \vee u_3 = 1 \end{cases} \\
x_4 &= \begin{cases} g_0(x_3) & \text{if } x_2 \vee u_4 = 0 \\ g_1(x_3) & \text{if } x_2 \vee u_4 = 1 \end{cases}
\end{aligned} \qquad (15)$$

where $u_1, u_2, u_3, u_4, x_1$, and $x_2$, are binary variables, $x_3, x_4 \in \{1, 2, 3, 4\}$, and the functions $g_0$ and $g_1$ are defined by

$$g_0(x) = \begin{cases} 2 & x \leq 3 \\ 3 & x = 4 \end{cases} \qquad g_1(x) = \begin{cases} 4 & x \geq 2 \\ 3 & x = 1 \end{cases}$$

It is not hard to see that $x_3$ and $x_4$ have a unique solution for every value of the $u$'s, and it is given by:

$$(x_3, x_4) = \begin{cases} (4,4) & \text{if } u_1 \vee u_3 = 1 \text{ and } u_2 \vee u_4 = 1 \\ (4,3) & \text{if } u_1 \vee u_3 = 1 \text{ and } u_2 \vee u_4 = 0 \\ (3,4) & \text{if } u_1 \vee u_3 = 0 \text{ and } u_2 \vee u_4 = 1 \\ (2,2) & \text{if } u_1 \vee u_3 = 0 \text{ and } u_2 \vee u_4 = 0 \end{cases}$$

The reason is that each of the four compositions: $g_0 g_0, g_1 g_1, g_0 g_1$, and $g_1 g_0$ has a unique fixed point given by 2, 4, 3 and 4, respectively. This is illustrated schematically by the graphs in Figure 3.

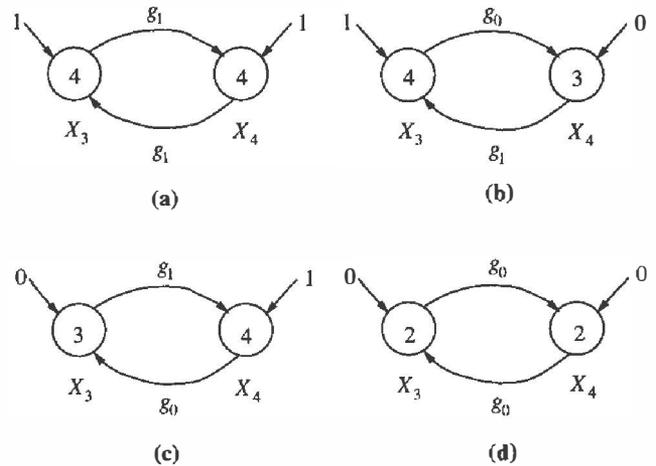

Figure 3: Showing the equilibrium solution of $X_3$ and $X_4$ for any combination of $U_1, U_2, U_3, U_4$. The labels at the incoming arrows correspond the values of to $u_1 \vee u_3$ and $u_2 \vee u_4$.

To verify that the independencies claimed by Eq. (13)–(14) hold in $P_T(x_1, x_2, x_3, x_4)$, we note that Eq. (13) follows from the fact the $X_1$ and $X_2$ are determined by two independent variables, $u_1$ and $u_2$. Eq. (14) follows from the fact (see Figure 3) that each state of $(x_3, x_4)$ dictates a unique value for $(u_3 \vee x_1, u_4 \vee x_2)$; thus, information about $x_1$ would not alter the probability of $x_2$, and vice versa.

Note that $(X_1 \perp X_2|X_3)$ and $(X_1 \perp X_2|X_4)$, though not reflected in $d$-separations, are also valid in $P_T$. This is perfectly consistent with the one-way implication in Eq. (8).

## 4  PROOFS OF THE MAIN RESULTS

We will prove Eq. (8) in three steps, each presented as a lemma.

**Lemma 3** *Let $G$ be any directed graph, possibly cyclic. For any disjoint sets of nodes $X, Y$ and $Z$ of $G$ we have*

$$(X \underline{\|}^* Y|Z)_{G^{XYZ}} \iff (G \underline{\|} Y|Z)_G \qquad (16)$$

**Proof:** Lemma 3 merely generalizes Eq. (3) to the case of cyclic graphs. Tracing the proof of Proposition 3 in [Lauritzen et al., 1990], we find that each step in the proof remains valid for cyclic graphs as well. Indeed, a "moralizing" edge is still added between two nodes, $a$ and $b$, only if $a$ and $b$ share some descendant in $Z$. Conversely, if $G$ contains a common descendant of $a$ and $b$ which is in $Z$, then there will be an edge between $a$ and $b$ in $G^{XYZ}$, either a moralizing edge or an original edge of $G$.

**Lemma 4** *Let $G_C$ be the undirected graph associated with a set $C$ of deterministic constraints on discrete variables $V = \{X_1, X_2, \ldots, X_n\}$, such that $G_C$ contains an edge between $X_i$ and $X_j$ iff there exists a constraint $c$ in $C$ that mentions both $X_i$ and $X_j$. Further, let $n(x), X \subseteq V$, stand for the number of solutions (of $C$) for which $X = x$. Then, for any three disjoint sets of variables $X, Y$, and $Z$, and for every instantiation $x, y$, and $z$, of $X, Y$, and $Z$, we have*

$$(X \underline{\|}^* Y|Z)_{G_C} \implies n(z)n(x,y,z) = n(x,z)n(y,z) \qquad (17)$$

**Proof:** Let $S_1$ be a subset of the variables including $X$ and $Z$, and $S_2$ a subset including $Y$ and $Z$ such that $V = S_1 \cup S_2$. (That $V$ can always be represented this way follows from the transitivity of undirected graphs [Pearl, 1988, Eq. (3.10e), p. 94]; every node outside $X \cup Y \cup Z$ must be separated by $Z$ from either $X$ or $Y$ (or both).) Denote by $n_i(x)$ the number of solutions in the set of all solutions projected on $S_i, i = 1, 2$. Since $Z$ separates $X$ from $Y$ and $S_1$ from $S_2$ in the graph $G_C$, we have

$$n(x,y,z) = n_1(x,z)n_2(y,z) \qquad (18)$$

Likewise, the separation of $Z$ implies

$$\begin{aligned} n(x,z) &= n_1(x,z)n_2(z) \\ n(y,z) &= n_2(y,z)n_1(z) \\ n(z) &= n_1(z)n_2(z) \end{aligned}$$

Substituting these three equalities into Eq. (18) gives Eq. (17). A special case of Lemma 4 was proved in [Dechter, 1990].

**Lemma 5** *Let $T$ be a semi-Markovian theory on a set $V$ of discrete variables, with associated probability function $P_T(v)$ and associated causal graph $G_T$. Then for every subset $W$ of variables, there exists a constraint problem $C_W$, with the same constraint graph $G_{C_W}$ as the moralized ancestral graph $G_T^W$ of $G_T$, such that*

$$P_T(w) = \alpha\, n_{C_W}(w) \qquad (19)$$

*where $\alpha$ is a constant, independent of $w$, and $n_{C_W}(w)$ is the number of solutions of $C_W$ that are compatible with $W = w$.*

**Proof:** We will prove Lemma 5 by constructing the desired constraint problem, $C_W$, for each set $W$ of variables.

First, we can assume without loss of generality that the variables in $U$ are discrete. This is legitimate because when $V$ is discrete, the domain of each $U_i$ can be partitioned into a finite number of equivalence classes, each containing those values of $U_i$ that are mapped (via the function $f_i$) into the same value of $X_i$.[3]

The next step in our construction is to replace the domain of each $U_i$ with a new, augmented domain, in which each value $u_i$ of $U_i$ is copied $K_i P(u_i)$ times, where $K_i$ is some large constant sufficient to make every $K_i P(u_i)$ term an integer.[4] By this augmentation, we have changed our theory $T$ to one in which all $U$ variables are discrete and uniformly distributed. The new theory, $T'$, has the same causal graph as $T$ and also induces the same probability $P_{T'(v)}$ on $V$. Moreover, $T'$ possesses the desirable feature that $P_{T'}(w)$ is equal to the fraction of solutions to a constraint problem $C$ made up of the functional constraints $\{f_i\}, i = 1, \ldots, n$, defined by the theory $T'$.

This constraint problem is close to fulfilling the conditions of Lemma 5, save for the fact that the constraint graph associated with $C$ is not $G_T$ but the fully moralized graph of $G_T$, namely $G_T^V$. At this point we invoke the fact that the constraints of $C$ are not arbitrary but are functional, namely, for every values of $pa_i$ and $u_i$ there is a solution for $x_i$. This implies that, for any set $W$ of variables, the equations associated with non-ancestors of $W$ do not constrain the permitted values of $W$ and can be omitted from the analysis (as if they were universal constraints). This completes the construction of $C_W$ as required by Lemma 5, because the constraint graph associated with $C_W$ is precisely the moralized ancestral graph of $G_T$, namely, $G_T^W$.

---

[3]These discrete $U$ variables were called *response function* variables in [Balke and Pearl, 1994] and *mapping* variables in [Heckerman and Shachter, 1995].

[4]We assume that $P(u_i)$ can be approximated by a rational number.

**Proof of Theorem 2**: Let $W$ be the union of $X, Y$, and $Z$. By virtue of Lemma 5, it is sufficient to prove that

$$(X \underline{\|}^* Y | Z)_{G_T^W} \Longrightarrow (X \perp Y | Z)_{P_T} \qquad (20)$$

The probabilistic independency on the r.h.s. of Eq. (20) amounts to the equality

$$P_T(z) P_T(x, y, z) = P_T(x, z) P_T(y, z) \qquad (21)$$

(see, for example, [Pearl, 1988, page 83]. Now, let $C_W$ be the constraint problem characterized in Lemma 5, for which we have

$$P_T(w) = \alpha\, n_{C_W}(w)$$

Moreover, this proportionality should hold for every subset $S$ of $W$ because, letting $S'$ stand for the variables in $W \backslash S$, we can write

$$\begin{aligned} P(s) &= \sum_{s'} P(s, s') = \alpha \sum_{s'} n(s, s') \\ &= \alpha\, n(s) \end{aligned}$$

by the definition of $n(s)$. Therefore, proving Eq. (20) amounts to proving

$$(X \underline{\|}^* Y | Z)_{G_T^W} \Longrightarrow n(z) n(x, y, z) = n(x, z) n(yz) \qquad (22)$$

According to Lemma 4 (Eq. (17)), the equality on the r.h.s. of Eq. (22) must hold in any constraint problem $C$ whose graph $G_C$ satisfies the separation $(X \underline{\|}^* Y | Z)_{G_C}$. But this separation is assured for $C_W$ by the antecedent of Eq. (22) and by the fact that $G_T^W$ coincides with the constraint graph of $C_W$.

## 5  CONCLUSION

We have shown that the $d$-separation criterion is valid for identifying independencies that result from causal mechanisms that include feedback provided that the variables are finite and discrete. Our finding should have direct application in the diagnosis of digital circuits and in programs that learn the structure of feedback systems [Richardson, 1996].

Spirtes (1994) has demonstrated, by a counterexample, that nonlinear continuous systems might violate the $d$-separation criterion when feedback is introduced. The results established in this paper mean that in continuous systems for which discrete-variable simulation gives a reasonable approximation, the impact of such violations is not too severe.

It should be noted, though, and this was hinted to us by an anonymous reviewer, that simulating continuous feedback systems with discrete variables is not a straightforward exercise. Simplistic attempts to replace each continuous function $f_i$ with a discrete version of $f_i$ would often end up with spurious instabilities, even when the original system is perfectly stable. The reason is that stability in feedback systems usually requires smooth, exponential approach toward an equilibrium point and such approach is ruled out by discretization, resulting in local traps in the forms limit cycles.

The example presented in section 3.1 avoids such traps by ensuring that the functions $f_3(f_4(\bullet, u_4), u_3)$ and $f_4(f_3(\bullet, u_3), u_4)$ each has a unique fixed point for each value of $u_3, u_4$. In general, to satisfy the unique-solution requirement of Definition 3, fixed-point conditions must be checked for every cycle in the system.

## Acknowledgments

The research of J. Pearl was supported by NSF grant #IRI-9420306, Air Force grant #F49620-93-1-0421, Rockwell/Northrop MICRO grant #95-118, and gifts from Microsoft Corporation and Hewlett-Packard Company. The research of R. Dechter was supported partially by NSF grant #IRI-9157636, Air Force grant #F49620-94-1-0173 and Rockwell International grant #UCM-20775.